\documentclass[runningheads]{llncs}
\usepackage{graphicx}
\usepackage{algpseudocode}
\usepackage{algorithm}
\usepackage[toc,page]{appendix}
\usepackage{float}
\graphicspath{ {./figures/} }

\begin{document}
\title{Refutation of Spectral Graph Theory Conjectures with Monte Carlo Search}
\titlerunning{Refutation of Spectral Graph Theory Conjecture with Monte Carlo Search}
%
\author{Milo Roucairol \and Tristan Cazenave}

%
\authorrunning{Roucairol and Cazenave} 
%
\institute{LAMSADE, Université Paris Dauphine - PSL, CNRS}
%
\maketitle              
\begin{abstract}
We demonstrate how Monte Carlo Search (MCS) algorithms, namely Nested Monte Carlo Search (NMCS) and Nested Rollout Policy Adaptation (NRPA), can be used to build graphs and find counter-examples to spectral graph theory conjectures in minutes.

\keywords{Monte Carlo Search \and Graph Theory \and Conjecture \and Refutation.}
\end{abstract}
\section{Introduction}

Monte Carlo search algorithm have proven to be powerful as game playing agents, with recent successes like AlphaGo\cite{holcomb_overview_2018}. These algorithms have the advantage of only needing an evaluation function for the final state of the space they explore. 

Graph conjectures are propositions on graph classes (any graph, trees, K-free...) that are suspected to be true and are awaiting a proof or a refutation. They lend themselves well to computer assisted proofs, as finding a counter example can be tedious to do manually. Spectral graph conjectures are appropriate to automated refutation because the property can often directly be turned into the evaluation function that can take many different values.
Thanks to software like Auto-GraphiX \cite{hansen_autographix_2000} and Graffiti \cite{delavina_history_nodate}, there are plenty of such conjectures.

Adam Zsolt Wagner showed that one could find explicit counter-examples using deep reinforcement learning of a policy with the deep cross entropy method \cite{wagner_constructions_2021}. 

In this paper, the MCS algorithms will play the game of refuting conjectures by building counter-examples.

First we will present the refutation of graph theory conjectures, then the different algorithms we use to explore the problem space, after that the procedure we use to build graphs and the game rules, finally we expose our results on four different conjectures.

\section{Refutation of Graph Theory Conjectures}

\subsection{Graph Theory Conjectures}

Conjectures in graph theory can be difficult to refute manually, unless one has an intuition of a counter-example, building a large number of graphs and computing invariant values or NP-hard problems on them often results in a waste of time. Hopefully, computers can help us with these score computations, the goal of automated conjecture refutation is to automatize the exploration as well.\\

Multiple types of graph theory conjectures exist: existence, topological, flow based, connectivity, cycle, minors, spectral...
The conjectures we are examining here are the spectral ones, they have the advantage of only requiring matrices calculations. The spectrum of a matrix is the set of its eigenvalues, spectral graph theory conjectures include conjectures on spectrum related invariants on different types of graph related matrices (adjacency, distance, laplacian...).\\

In this article we aim at evaluating the interest of Monte Carlo Search methods in order to refute some spectral graph theory conjectures. We are interested in conjectures that Adam Zsolt Wagner refuted in his article \cite{wagner_constructions_2021} as well as in other conjectures.

\subsection{Algorithms Used to Refute Graph Conjectures}

Wagner used deep neural networks with cross entropy. The network is used to learn a policy from a state, it is trained by constructing a batch of graphs according to the policy, evaluating and selecting the best graphs from the batch to adjust the neural network with their state-moves couples according to their scores (obtained during the evaluation), then it's repeated starting by the generation of a new batch until the conjecture is refuted.\\

Some of our conjectures come from Aouchiche and Hansen survey \cite{aouchiche_survey_2010}. In order to explain how Graffiti selects conjectures they state:\\
“Graffiti generates many conjectures of a simple form (e.g. inequalities between two invariants or
between an invariant and the sum of two others) then tests them on a database of graphs and discards
those which are falsified. Should this test be passed, it is checked if the formulas are implied by known
ones (in which case they are also discarded) and that they provide new information for at least one
graph in the database, i.e., that they are stronger than the conjunction of all other formulas for that
graph. If not, they are temporarily set aside. If yes, they are proposed to all graph theorists, in the
electronic file “Written on the Wall”, which reports on the status of almost 1000 conjectures.
Many well-known graph theorists worked on these conjectures and this led to several dozen papers.
Some of Graffiti’s conjectures are about various topics in spectral graph theory, namely, the eigenvalues,
as well as their multiplicity, of the adjacency, Laplacian and distance matrices of graphs.”

\section{Algorithms}

In this section we present  Monte Carlo Search and the different search algorithms we use: NMCS, NRPA and Greedy-BFS.

\subsection{Monte Carlo Search}

Monte Carlo Tree Search (MCTS) has been successfully applied to many games and problems \cite{BrownePWLCRTPSC2012}.

Nested Monte Carlo Search (NMCS) \cite{CazenaveIJCAI09} is an algorithm that works well for puzzles and optimization problems. It biases its playouts using lower level playouts. At level zero NMCS adopts a uniform random playout policy. Online learning of playout strategies combined with NMCS has given good results on optimization problems \cite{RimmelEvo11}. Other applications of NMCS include Single Player General Game Playing \cite{Mehat2010}, Cooperative Pathfinding \cite{Bouzy13}, Software testing \cite{PouldingF14}, heuristic Model-Checking \cite{PouldingF15}, the Pancake problem \cite{Bouzy16}, Games \cite{CazenaveSST16} and the RNA inverse folding problem \cite{portela2018unexpectedly}.

Online learning of a playout policy in the context of nested searches has been further developed for puzzles and optimization with Nested Rollout Policy Adaptation (NRPA) \cite{Rosin2011}. NRPA has found new world records in Morpion Solitaire and crosswords puzzles. NRPA has been applied to multiple problems: the Traveling Salesman
with Time Windows (TSPTW) problem \cite{cazenave2012tsptw,edelkamp2013algorithm}, 3D Packing with Object Orientation \cite{edelkamp2014monte}, the physical traveling salesman problem \cite{edelkamp2014solving}, the Multiple Sequence Alignment problem \cite{edelkamp2015monte} or Logistics \cite{edelkamp2016monte}. The principle of NRPA is to adapt the playout policy so as to learn the best sequence of moves found so far at each level.

The use of Gibbs sampling in Monte Carlo Tree Search dates back to the general game player Cadia Player and its MAST playout policy \cite{finnsson2008simulation}.

\subsection{Nested Monte Carlo Search}

NMCS \cite{CazenaveIJCAI09} is a Monte Carlo Search algorithm that recursively calls lower level NMCS on children states of the current state in order to decide which move to play next, the lowest level of NMCS being a random playout, selecting uniformly the move to execute among the possible moves. A heuristic can be added to the playout choices, but it is not the case with the NMCS used here. \\

Algorithm \ref{NMCS} gives the NMCS algorithm. The different notations used are:\\

$M_{current\mbox -state}$ denotes the legal moves available from the state $current\mbox -state$.\\

$randomChoice(L)$ is a function that returns an element selected uniformly from $L$.

\subsection{Nested Rollout Policy Adaptation}

Introduced by Christopher D. Rosin \cite{Rosin2011}, NRPA is akin to mixing NMCS \cite{CazenaveIJCAI09} and Q-learning \cite{watkins_q-learning_1992}, performances are generally better than NMCS but the exploration/exploitation can lock itself in a local minimum. The algorithm is also dependent on how moves are defined.

The algorithm uses a policy which consist in attributing a value to a move. The difference with Wagner's policy \cite{wagner_constructions_2021} is here the value of the move is not necessarily learned given the state of the graph, but given the definition of the move, the policy can thus be defined given the few previous moves or the number of moves preceding the current one. It is also a key difference between NRPA policy learning and Q-learning, Wagner's policy is closer to deep Q-learning.

Like the NMCS, the NRPA calls other lower level NRPA which start with the higher level current policy, these lower level NRPA return their best sequence which is used to update the higher level policy. At the lowest level, the NRPA launches a playout and randomly select moves according to the policy.

$softmaxChoice(L, policy)$ is a function that returns an element selected according to a softmax distribution on each element value mapped on the policy.
Chances of selecting $e$ from $L$ are $\frac {exp(policy[e])}{sum([exp(policy[i]) for \; i \; in \; L])}$.\\

In all our experiments, NRPA's $Alpha$, the learning rate, is set to 1.\\

Algorithm \ref{NRPA} gives the NRPA algorithm.

\subsection{Greedy Best First Search}

Another algorithm we use is Greedy-BFS \cite{doran1966experiments}. It consists in opening the best (or randomly one of the best) previously evaluated node, evaluating all its children and inserting these children in a list, sorted by their evaluation scores.

The evaluation can be done with a heuristic (in which case it is not Monte Carlo) or a Monte Carlo algorithm like a playout or a low level NMCS.

In this paper, the Greedy-BFS was used as a general search technique and not a Monte Carlo search algorithm. We use the score function (or the modified evaluation function in conjecture 2) to choose which leaves to expand.

\section{Graph Generation}

For each problem we want to solve, we define an interface. This interface provides, to the search algorithm, the classes for the $state$ and for the $move$ and their associated functions: $score$, $terminal$, $M_{state}$ and $play$.

The goal is to find an instance of a class of graphs that does not respect a property speculated on that class of graphs, in order to do so we use reinforcement learning and tree search algorithms to explore the possible graphs from that class and hopefully converge to a counter-example.

Moves to generate graphs are the same in all the approaches presented here, the graph is built edge by edge. Moves are represented as couple of integers corresponding to the two vertices the edge will link. If one of the two integer corresponding vertex is not in the graph yet, it is added alongside the edge as a leaf (-1 as the second member of the couple means adding a leaf anyway).\\

What differs from one conjecture (interface) to another are the methods :\\
- $score$ which derives from the conjecture we want to refute.\\
- $terminal$ is necessary to MCS methods to know when stop a playout, it can be the size of the graph (what we used here) but also the number of edges or any property on the graph. Trying different constraints for that method is part of the experiments.\\
- $M_{state}$ gives the legal moves from a state according to the model. For example a model of maximum degree 3 will exclude any moves that result in the edge addition on a vertex of degree 3.

\section{Experiments on Conjectures}

The experiments were made with Rust 1.59, on a Intel Core i5-6600K 3.50GHz using a single core (but parallel processing is very accessible).

Every solution was verified using WolframAlpha symbolic computing of eigenvalues to ensure they were not due to floating point errors in Rust nalgebra 0.30.1 library.

We express every score function in order to maximise them.

\subsection{Conjecture 1. AutoGraphiX}

Conjecture 7 from \cite{aouchiche_survey_2010}, also present as conjecture 2.1 from \cite{wagner_constructions_2021}, is used as a proof of concept in the aforementioned paper. We decided to refute all conjectures from \cite{wagner_constructions_2021} to show that MCS can be equally as good as the deep cross entropy.\\

\textbf{Definition 1.} \textit{Let G be a connected graph, the matching number is the size of the matching that contains the largest number of edges of G}.\\

\textbf{Definition 2.} \textit{Let G be a connected graph on $n$ vertices, the distance spectrum $\lambda_1, ..., \lambda_n$ is the spectrum of the distance matrix, ranked in descending order ($\lambda_1$ being the largest, the index)}.\\

This conjecture is the following :

\textbf{Conjecture 1.} \textit{Let G be a connected graph on $n \geq 3$ vertices with index $\lambda_1$ and matching number $\mu$. Then $\lambda_1 + \mu \geq \sqrt{n-1} + 1$}.\\

$\lambda_1 + \mu \geq \sqrt{n-1} + 1  \Leftrightarrow 0 \geq \sqrt{n-1} + 1 - \lambda_1 - \mu$\\

Which gives \\

\textbf{Score function :} $\sqrt{n-1} + 1 - \lambda_1 - \mu$ \\

\begin{figure}[H]
    \centering
    \includegraphics[width=8cm]{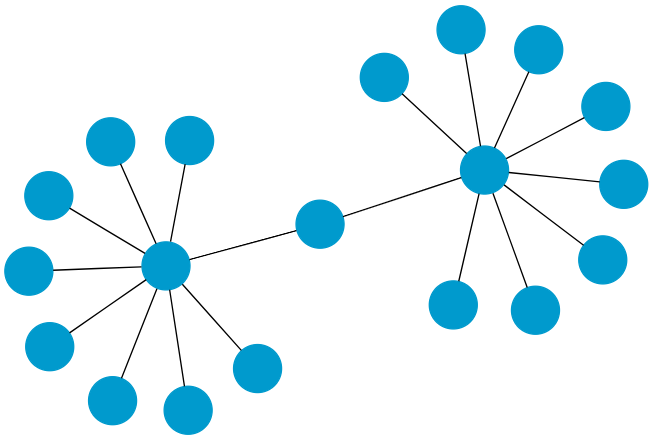}
    \caption{Counter example of conjecture 1 of size 19}
    \label{conj1_0}

\end{figure}

With a NRPA of level 3, and generating trees of size 19, we found a graph with a positive score of 0.016: $\lambda_1 + \mu \approx 5.227 $ and $ \sqrt{n-1} + 1 \approx 5.243$ \\
It is the same counter-example (figure \ref{conj1_0}) as Adam Zsolt Wagner \cite{wagner_constructions_2021}. His results helped us as we knew what to search for and that we could exploit NRPA policy learning to recreate the pattern found in his counter example. However it is interesting to try generating trees first anyway when refuting a spectral graph conjecture as the space to explore is way smaller than for any graph, and trees are extreme instances of the class, so these tend to be counter-examples. 

The NRPA algorithm can lock itself in a policy leading to a local minimum while searching a counter-example to that conjecture. It is important to use restarts in this case: if a result is not found in less than a second the algorithm is launched again a few number of times (usually 10 to 20 times) until the refutation is found. While it takes a few hours to find the counter example with Wagner's deep cross entropy \cite{wagner_constructions_2021}, our method can find it in seconds even with multiple relaunches. Using restarts this way can be parallelized easily.

\begin{figure}[H]
    \centering
    \includegraphics[width=8cm]{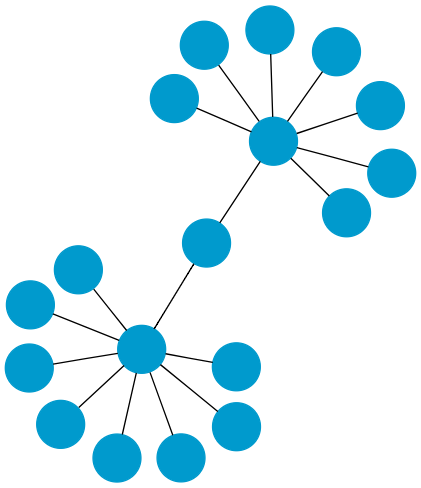}
    \caption{Counter example of conjecture 1 of size 18}
    \label{conj1_1}

\end{figure}

We also were able to find a counterexample of size 18 (figure \ref{conj1_1}) with the same process, with a positive score of 0.012: $\lambda_1 + \mu \approx 5.101 $ and $ \sqrt{n-1} + 1 \approx 5.123$ \\

\subsection{Conjecture 2. Aouchiche-Hansen}

Conjecture 2.15 from \cite{aouchiche_proximity_2016}, also present as conjecture 2.3 from \cite{wagner_constructions_2021}.\\

\textbf{Definition 3.} \textit{Let G be a connected graph, the diameter is the maximum length of the shortest paths between all vertices of G}.\\

\textbf{Definition 4.} \textit{Let G be a connected graph, the proximity is, over all the vertices of G, the minimum average distance to all the other vertices of G}.\\

\textbf{Conjecture 2.} \textit{Let G be a connected graph on $n \geq 4$ vertices with diameter $D$, proximity $\pi$ and distance spectrum $\lambda_1 \geq ... \geq \lambda_n$. Then $\pi + \lambda_{ \lfloor \frac{2D}{3} \rfloor} \geq 0$}.\\

Results from Wagner paper helped us to know what kind of graph generate: a tree with more than 203 vertices. The intuition of that counter-example can also be seen in \cite{aouchiche_proximity_2016}.

The NRPA algorithm, with the standard score function, finds trees of size 30 with similar scores (-0.4) as the deep cross entropy method \cite{wagner_constructions_2021} in 3s, when it takes days of training to the deep cross entropy to achieve these scores.

However none of the Monte Carlo method found a counter example with these information only, just like Wagner's deep cross entropy.
We determined that the problem stemmed from the score function which lumps scores from different graphs close together, a graph can see its diameter increase but not see repercussions on its score due to the flooring (increasing the diameter of the graph is indeed the way to find a counter example).
To solve this problem, we designed an evaluation function slightly different from the score function:\\

\textbf{Score function:} $-\pi - \lambda_{ \lfloor \frac{2D}{3} \rfloor}$ \\

\textbf{Evaluation function:} $-\pi - \lambda'_ {2D} $ \\

We define $\lambda'$ by linearly interpolating $\lambda$ to be 3 times as long.\\

\begin{figure}[H]
    \centering
    \includegraphics[angle = 90, width=12cm]{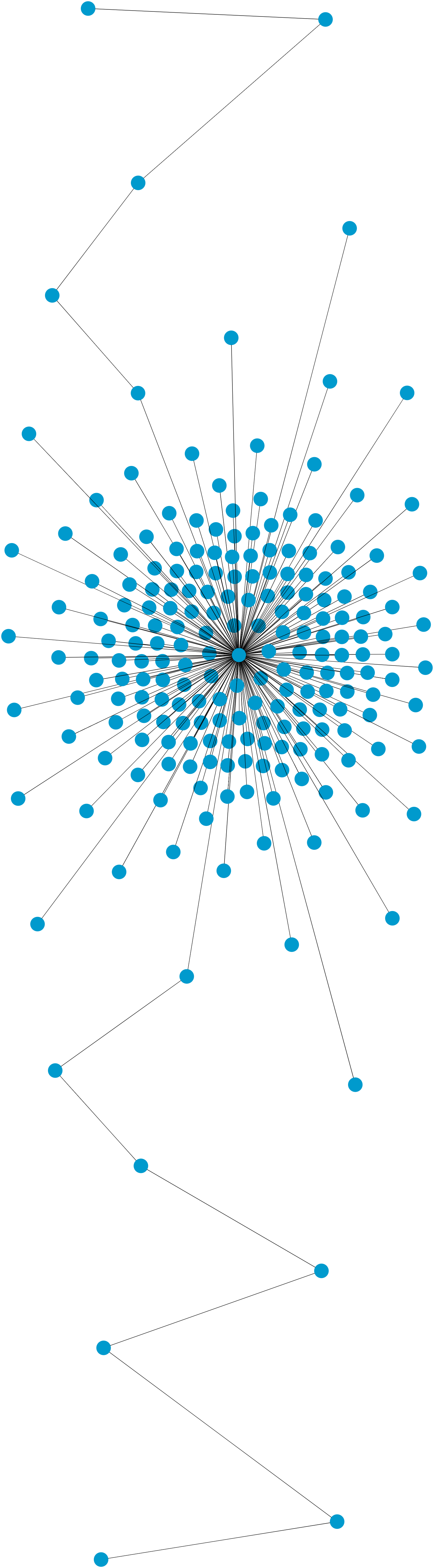}
    \caption{Counter example of conjecture 2}
    \label{conj2}

\end{figure}

With this new evaluation function, increasing the diameter of the graph always leads to a very different score. Instead of playouts, we then used that evaluation function in the greedy-BFS algorithm which found a counter example (figure \ref{conj2}) in about 5 minutes.\\

The graph shown in figure \ref{conj2} had a score of approximately 0.000285, meaning $\pi + \lambda_{ \lfloor \frac{2D}{3} \rfloor} < 0$, it's indeed a counter-example (it was verified to not be a floating point error with wolfram symbolic eigenvalues computing).

\subsection{Conjecture 3. Collins}

Conjecture 10 from \cite{collins_conjecture_1989}, also present as conjecture 2.4 from \cite{wagner_constructions_2021}.

Given a tree $T$ on n vertices, its adjacency matrix $A(T)$ and its distance matrix $D(T)$.\\

$CPD(T)$ is the characteristic polynomial of $D(T)$: \\

$$CPD(T) = det(D(T)-Ix) = \sum_{k = 0}^{n} \delta_k x^k $$ \\

Let the coefficients $d_k = \frac{2^k}{2^{n-2}} |\delta_k|$ for k in {0, ..., n-2} be the normalized coefficient of the characteristic polynomial.\\

Let $p_{D}(T)$ be the peak of the normalized coefficient of the characteristic polynomial.\\

$CPA(T)$ is the characteristic polynomial of $A(T)$: \\

$$CPA(T) = det(A(T)-Ix) = \sum_{k = 0}^{n} a_k x^k$$

Let $p_{A}(T)$ be the peak of the non-zero coefficients ($a_k$) of $CPA(T)$.\\

\textbf{Conjecture 3.} \textit{Given a tree T, $CPA(T)$ form an unimodal sequence and its peak $p_{A}(T)$ is at the same place as $p_{D}(T)$ }.\\

Like Adam Zsolt Wagner \cite{wagner_constructions_2021}, we only refuted the second part of the conjecture with a level 2 NMCS, using the same score function. \\

\textbf{Score function :} $|\frac{p_{A}(T)}{\# \lambda_{\ne 0}} - (1 - \frac{p_{D}(T)}{n-2})|$ \\

\begin{figure}[H]
    \centering
    \includegraphics[angle = 90,width=12cm]{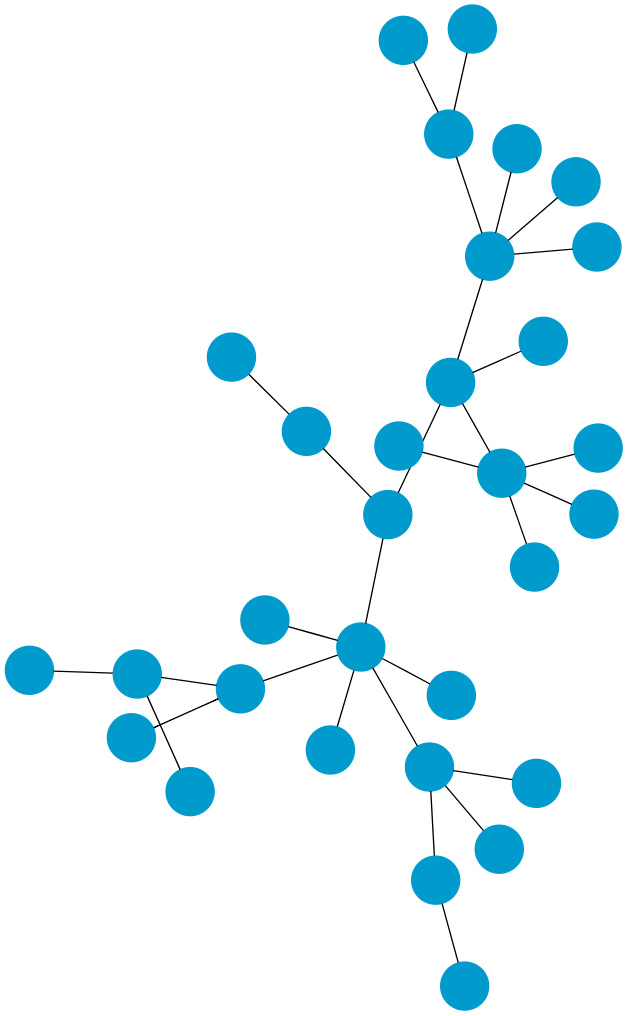}
    \caption{Counter example of conjecture 3}
    \label{conj3}

\end{figure}

The counter-example in figure \ref{conj3} was found in less than a second, it features a $p_{A}(T)$ at 19/32 and a $p_{D}(T)$ at 16/30 which gives no doubt that the peaks are at different positions.

\subsection{Conjecture 4. Graffiti 137}

That conjecture was made automatically by a program named Graffiti \cite{delavina_history_nodate}, it is present in the survey \cite{aouchiche_survey_2010} as an already refuted conjecture.\\

Let $Hc = \sum_{uv \in E}^{} \frac{1}{d(u) + d(v)}$ the harmonic of graph $G$ with vertices $E$, $d(u)$ is the degree of the vertex $u$.\\

\textbf{Conjecture 4.} \textit{For any graph G, the second biggest adjacency matrix eigenvalue $\lambda_2$ is inferior to the harmonic of the graph : $\lambda_2 < Hc(G)$}.\\

\textbf{Score function :} $\lambda_2 - Hc(G)$ \\

\begin{figure}[H]
    \centering
    \includegraphics[width=4cm]{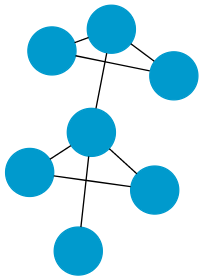}
    \caption{Counter example of conjecture 4}
    \label{conj5}

\end{figure}

A counter-example of size 7 (figure \ref{conj5}) is quickly found by NMCS of level 2, NRPA of level 2 or even Greedy-BFS with $\lambda_2 \approx 1.786 $ and $ Hc(G) = 5/3 \approx 1.576 $ .\\

\subsection{Comparison of Search Algorithms}

Table 1 gives the times required to refute each conjecture for each algorithm. The machine used has a i5-6600K 3.5 GHz CPU. We see that NMCS can refute conjectures 3 and 4 almost instantly but does not find refutations to conjectures 1 and 2. NRPA also refutes conjectures 3 and 5 almost instantly and refutes conjecture 1 in 0.1 seconds when Wagner solves it in a few hours. Greedy-BFS refutes conjectures 2 and 4. It is the only algorithm (among those we tried) able to refute conjecture 2, using a modified score function to guide a non MC search. It solves it in 5 minutes when Wagner algorithm solves it in a few days. 

\begin{table}[h!]
\caption{Summary of our results}
\label{table:1}
\begin{center}
\begin{tabular}{||c | c c c c||} 
 \hline
 Method & Conj 1 & Conj 2* & Conj 3 & Conj 4 \\ [1.5ex] 
 \hline\hline
 NMCS & - & - & 1s (lv 2) & 0s (lv 2) \\ [1.5ex] 
 \hline
 NRPA & 1s (lv 3) & - & 1s (lv 2) & 0s (lv 2) \\ [1.5ex] 
 \hline
 Greedy-BFS & - & 291s & - & 0s \\ [2ex] 
 \hline
\end{tabular}
\end{center}
(-) denotes a failure, an inability to refute the conjecture.

* Refuted using the evaluation function, different from the score function, that evaluates non terminal states.\\
\end{table}

Conjecture 2 could not be refuted with the natural score function derived from the conjecture but could be solved with Greedy-BFS using a more informative score function. NRPA of level 2 with the natural score function could attain a score of -0.4 in 3s instead of a few days in Wagner's work \cite{wagner_constructions_2021}.

\section{Conclusion}

Monte Carlo search methods proved to be powerful ways of refuting conjecture from spectral graph theory much faster than Wagner's deep cross entropy method \cite{wagner_constructions_2021}. Trying to build trees even when the conjecture is applied on any graph can also be helpful as it reduces the amount of possible builds greatly, it is inexpensive and should be tried first.

However, these methods present limits. Computing score functions that require eigenvalues on big trees (over size 500) can be very costly. They are also dependent on the shape of the score function: a noisy score function with many local minimum can be challenging, as well as a score function with more discrete results can lead to an absence of differentiation in the paths to explore (see conjecture 2). Conjectures requiring to compute a NP hard problem can also severely increase the computing time even for small graphs (30 vertices).

In the future we aim to refute more conjectures and to improve the Greedy-BFS with MCS method, potentially with pruning strategies too.

\bibliographystyle{splncs04}
\bibliography{referencesMod}

\newpage
\appendix

\section{Appendix - Algorithms}

\begin{algorithm}[H]
  \begin{algorithmic}
    \State{NMCS ($current\mbox -state$, $level$)
    \If{$level = 0$}
     \State{$ply \leftarrow 0$}
     \State{$seq \leftarrow \{\}$ }
     \While{$current\mbox -state$ is not terminal}
      \State{$move \leftarrow randomChoice(M_{current\mbox -state})$}
      \State{$current\mbox -state \leftarrow play(current\mbox -state, move)$ }
      \State{$seq[ply] \leftarrow move$}
      \State{$ply += 1$}
     \EndWhile
     \State{}
     \Return{score ($current\mbox -state, seq$)}
    \Else
     \State{$best\mbox -score \leftarrow - \infty$}
    \While{$current\mbox -state$ is not terminal}
    \For{{\bf each} $move$ in $M_{current\mbox -state}$}
     \State{$next\mbox -state \leftarrow play (current\mbox -state, move)$}
     \State{$(score, seq) \leftarrow $NMCS ($next\mbox -state, level-1) $}
    \If{$score \ge best\mbox -score$}
     \State{$next\mbox -best\mbox -state \leftarrow next\mbox -state$}
     \State{$best\mbox -score \leftarrow score$}
     \State{$best\mbox -sequence \leftarrow seq$}
    \EndIf
      \EndFor
      \State{$current\mbox -state \leftarrow next\mbox -best\mbox -state$}
       \EndWhile
       \State{}
       \Return{($best\mbox -score, best\mbox -sequence$)}
  \EndIf}
\end{algorithmic}
\caption{\label{NMCS}The NMCS algorithm.}
\end{algorithm}

\begin{algorithm}[H]
  \begin{algorithmic}
    \State{NRPA ($policy$, $level$)
    \If{$level = 0$}
     \State{$current\mbox -state \leftarrow root()$}
     \State{$ply \leftarrow 0$}
     \State{$seq \leftarrow \{\}$ }
     \While{$current\mbox -state$ is not terminal}
      \State{$move \leftarrow softmaxChoice(M_{current\mbox -state}, policy)$}
      \State{$current\mbox -state \leftarrow play(current\mbox -state, move)$ }
      \State{$seq[ply] \leftarrow move$}
      \State{$ply += 1$}
     \EndWhile
     \State{}
     \Return{ (score ($current\mbox -state$), $seq$)}
    \Else
     \State{$best\mbox -score \leftarrow - \infty$}
     \For{ $N$ $iterations$}
        \State{$(result, new) \leftarrow$ NRPA($policy$, $level-1$)}
        \If{$result \geq best\mbox -score$}
            \State{$best\mbox -score \leftarrow result$}
            \State{$seq \leftarrow new$}
        \EndIf
        \State{$pol \leftarrow adapt(pol, seq)$}
     \EndFor
     \State{}
     \Return{($best\mbox -score, seq$)}
  \EndIf}
\end{algorithmic}
\medskip{}
\begin{algorithmic}
  \State{Adapt ($policy$, $seq$)
  \State{$node \leftarrow root()$}
  \State{$pol'\leftarrow  pol$}
  \For{$ply = 0$ TO $length(seq)-1$}
    \State{$pol'[(node, seq[ply])]$ +=  $Alpha$}
    \State{$z \leftarrow  Sum([exp(pol[(node, m)])$ for $m$ in $M_{node} ])$}
    \For{{\bf each} $move$ in $M_{node}$}
        \State{$pol'[(node, move)]$ -= $\frac{Alpha \cdot exp(pol[(node, move)])}{z}$}
    \EndFor
    \State{$node \leftarrow play(node, seq[ply])$}
  \EndFor\\
  \Return{$pol'$}}

\end{algorithmic}
\caption{\label{NRPA}The NRPA algorithm.}
\end{algorithm}

\section{Appendix - Graphs}

    Edges:
0-1, 0-2, 0-3, 0-4, 0-5, 0-6, 0-9, 0-11, 0-18, 4-7, 7-8, 7-10, 7-12, 7-13, 7-14, 7-15, 7-16, 7-17

    Edges:
0-1, 0-2, 0-3, 0-4, 0-5, 0-6, 0-9, 0-11, 4-7, 7-8, 7-10, 7-12, 7-13, 7-14, 7-15, 7-16, 7-17

    Edges:
0-1, 0-8, 0-13, 0-14, 0-15, 0-16, 0-17, 0-18, 0-19, 0-20, 0-21, 0-22, 0-23, 0-24, 0-25, 0-26, 0-27, 0-28, 0-29, 0-30, 0-31, 0-32, 0-33, 0-34, 0-35, 0-36, 0-37, 0-38, 0-39, 0-40, 0-41, 0-42, 0-43, 0-44, 0-45, 0-46, 0-47, 0-48, 0-49, 0-50, 0-51, 0-52, 0-53, 0-54, 0-55, 0-56, 0-57, 0-58, 0-59, 0-60, 0-61, 0-62, 0-63, 0-64, 0-65, 0-66, 0-67, 0-68, 0-69, 0-70, 0-71, 0-72, 0-73, 0-74, 0-75, 0-76, 0-77, 0-78, 0-79, 0-80, 0-81, 0-82, 0-83, 0-84, 0-85, 0-86, 0-87, 0-88, 0-89, 0-90, 0-91, 0-92, 0-93, 0-94, 0-95, 0-96, 0-97, 0-98, 0-99, 0-100, 0-101, 0-102, 0-103, 0-104, 0-105, 0-106, 0-107, 0-108, 0-109, 0-110, 0-111, 0-112, 0-113, 0-114, 0-115, 0-116, 0-117, 0-118, 0-119, 0-120, 0-121, 0-122, 0-123, 0-124, 0-125, 0-126, 0-127, 0-128, 0-129, 0-130, 0-131, 0-132, 0-133, 0-134, 0-135, 0-136, 0-137, 0-138, 0-139, 0-140, 0-141, 0-142, 0-143, 0-144, 0-145, 0-146, 0-147, 0-148, 0-149, 0-150, 0-151, 0-152, 0-153, 0-154, 0-155, 0-156, 0-157, 0-158, 0-159, 0-160, 0-161, 0-162, 0-163, 0-164, 0-165, 0-166, 0-167, 0-168, 0-169, 0-170, 0-171, 0-172, 0-173, 0-174, 0-175, 0-176, 0-177, 0-178, 0-179, 0-180, 0-181, 0-182, 0-183, 0-184, 0-185, 0-186, 0-187, 0-188, 0-189, 0-190, 0-191, 0-192, 0-193, 0-194, 0-195, 0-196, 0-197, 0-198, 0-199, 0-200, 0-201, 0-202, 1-2, 2-3, 3-4, 4-5, 5-6, 6-7, 8-9, 9-10, 10-11, 11-12

    Edges:
0-1, 0-2, 0-13, 0-27, 1-5, 1-7, 1-9, 1-22, 2-3, 3-4, 3-8, 3-12, 4-10, 4-25, 5-6, 5-16, 5-21, 6-14, 6-28, 9-11, 10-20, 11-24, 13-19, 14-15, 15-17, 15-18, 15-23, 19-26, 19-30, 28-29

    Edges:
0-1, 0-2, 1-2, 0-3, 3-4, 4-5, 3-5, 3-7

\end{document}